\documentclass[11pt]{article}

\usepackage[final]{acl}

\usepackage{times}
\usepackage{latexsym}

\usepackage{amsmath}
\usepackage{amssymb}
\usepackage{bm}
\usepackage{mathtools}
\usepackage{multirow}
\usepackage{booktabs}
\usepackage{array}
\usepackage{adjustbox}
\usepackage{subcaption}
\usepackage{colortbl}
\usepackage{xcolor}
\usepackage{pifont}
\usepackage{enumitem}

\newcommand{\cmark}{\ding{51}}

\definecolor{rowhl}{HTML}{F2F2F2}
\newcommand{\Apos}{\hat{A}^{+}}
\newcommand{\Aneg}{\hat{A}^{-}}
\newcommand{\npos}{n^{+}}
\newcommand{\nneg}{n^{-}}
\usepackage[T1]{fontenc}

\usepackage[utf8]{inputenc}

\usepackage{microtype}

\usepackage{inconsolata}

\usepackage{graphicx}
\usepackage{algorithm}
\usepackage{algpseudocode}

\begin{document}

\title{Spurious Advantage Hidden in GRPO}

\author{
  \textbf{Jiamian Wang\textsuperscript{1}},
  \textbf{Samyadeep Basu\textsuperscript{2}},
  \textbf{Koustava Goswami\textsuperscript{2}},
  \textbf{Tong Yu\textsuperscript{2}},
  \textbf{Zhiqiang Tao\textsuperscript{1}}
\\
  \textsuperscript {1}Rochester Institute of Technology,
  \textsuperscript {2}Adobe Research
}

\maketitle
\begin{abstract}
Group Relative Policy Optimization (GRPO) is widely studied for reinforcement learning with verifiable rewards, where its advantage estimator assigns each rollout a magnitude from within-group reward statistics. In the common case, this magnitude rewards rollouts that reach the correct answer through reasoning. Yet, an overlooked case shares the same surface: a rollout may land on it by guessing, and the formula still assigns a high magnitude, which we identify as the \emph{spurious advantage}. This arises in three cases: bounded-answer tasks with a small candidate set; open-answer sets hosting bounded sub-cases; and search agents whose budget opens many paths to the same answer. In all three, this misleads the policy toward guess-like behaviors. We propose \textsc{SignBalance}, whose magnitude is composition-free: it keeps the verifier sign, uses a global scale, and restores zero-mean balance via a stop-gradient per-class rescaling. Across math and search-agent benchmarks at different scales, \textsc{SignBalance} matches GRPO on open-answer math and improves on bounded-answer math and search agents. Code will be released. 
\end{abstract}

\section{Introduction}\label{sec:intro}

Reinforcement learning with verifiable rewards (RLVR) has become the standard recipe for post-training large language models on reasoning tasks, and Group Relative Policy Optimization (GRPO)~\cite{shao2024deepseekmath} is  widely studied in this line, where a group of rollouts is sampled per prompt, each rollout is scored by a binary verifier, and the per-rollout advantage is constructed from within-group reward statistics~\cite{shao2024deepseekmath,guo2025deepseek}. Recent improvements operate around this estimator, \emph{e.g.}, refining the clip range, replacing or removing the normalization term, reshaping the sampling strategy, \emph{etc.}~\cite{yu2026dapo,liu2025understanding,xiao2025bnpo,wu2025quantile}.

The GRPO advantage estimator applies the same within-group normalization formula on every task, where the magnitude assigned to each rollout is determined by the in-group correct/wrong composition, regardless of whether the prompt is an open-answer math problem or a multi-choice question, and whether the rollout is a single answer or a multi-turn search trajectory. In the common case, a correct rollout earns its magnitude because the policy reasoned through the problem, which serves as an effective weight for the gradient. \emph{Yet, an overlooked case shares the same surface}: a rollout can land on the correct answer by guessing one candidate answer rather than by reasoning, still attaining it a high magnitude. The estimator does not distinguish the two. The per-rollout gradient weight then contains a component that comes from guessing, misleading the policy toward learning lucky-guess trajectories instead of reasoning. We call this component the \emph{spurious advantage}.

The spurious advantage shows up most directly on bounded-answer tasks. Consider a multi-choice prompt with $k$ options (e.g., $k=4$): even without any reasoning, a policy that randomly picks one option still hits the gold answer with probability $1/k$. Within a group of $G$ rollouts, a rollout that arrives at the correct option by guessing and another that arrives at it by reasoning receive the very same per-rollout magnitude from the within-group formula, and therefore feed the policy update with the same gradient weight. The guessing one is a noisy signal that latently encourages the policy to imitate lucky guesses rather than to reason.

Interestingly, this phenomenon is not exclusive to bounded-answer benchmarks; it can quietly extend to other settings whose effective answer surface remains finite. On the training-data side, even a nominally open-answer math corpus can be far from uniformly open: in MATH-7.5K, $55.95\%$ of problems admit short numeric or categorical answers, so the corpus is in fact a mixture of many small bounded sub-cases (Table~\ref{tab:math75k-category}), leaving the policy a non-negligible probability of guessing into the gold answer. On the search-agent side, a multi-turn rollout with a larger action budget explores a wider set of possible paths to the final answer, and many of these paths involve redundant or invalid searches yet still cash out as correct via outcome-only reward. The estimator assigns those guessed or detoured rollouts the same high magnitude as the genuinely reasoned ones, so the spurious advantage enters the gradient weight in both settings. This motivates an advantage estimator that separates the guessing component from the reasoning one.

To this end, we propose an advantage estimator, \textsc{SignBalance}, whose per-rollout magnitude does not depend on the within-group composition, \emph{i.e.}, $(\npos, \nneg)$. The spurious component no longer enters the per-rollout gradient weight. \textsc{SignBalance} keeps the verifier sign of each rollout, replaces the composition-dependent magnitude with a global scale, and restores the batch-level zero-mean force balance through a stop-gradient per-class rescaling. It is parameter-free, leaves the PPO-style surrogate unchanged, and adds no external model or inference-time cost. 
We summarize the contributions as follows:
\begin{itemize}[nosep,leftmargin=1.5em]
    \item This work isolates a  \emph{spurious advantage} phenomenon in  GRPO advantage estimator, which can mislead the policy toward learning guess-like behaviors rather than effective reasoning.
    \item This work identifies three cases where this component becomes large: bounded-answer tasks, bounded cases hidden in open-answer training sets, and multi-turn search agents with outcome-only reward, providing a unified lens to revisit GRPO's learning signal.
    \item This work proposes \textsc{SignBalance}, an advantage estimator whose magnitude does not depend on within-group composition,  is parameter-free, keeps the PPO-style surrogate unchanged, and introduces no external model or inference-time cost.
    \item Experiments across math reasoning and search-agent tasks, and across different model scales, show consistent gains.
We hope this work may invite further attention to how GRPO interacts with the structure of specific tasks.
\end{itemize}

\section{Related Work}\label{sec:related}

\textbf{GRPO and Its Variants.}
GRPO~\cite{shao2024deepseekmath} estimates per-rollout advantage from the in-group reward distribution and removes the PPO value network, and has driven strong results on open-answer math reasoning at scale~\cite{shao2024deepseekmath,guo2025deepseek}. A rich body of follow-up work has refined the surrounding mechanics on clipping and sampling~\cite{yu2026dapo,xiong2025reinforce}, on normalization and aggregation~\cite{liu2025understanding,xiao2025bnpo,wu2025quantile,zhao2025geometric,he2025deltal,jin2025tic}, on advantage shaping~\cite{wang2025lambda,huang2025mapo,chen2025conditional,zhang2025edge}, on the treatment of wrong rollouts and all-wrong groups~\cite{zhu2026surprising,nan2025ngrpo,feng2025don,li2025drpo}, and on multi-objective composite rewards~\cite{ichihara2025mo}. These efforts achieve strong empirical results within their respective scopes; their evaluations and design choices center on open-answer math, where the GRPO magnitude assignment can be treated as a universally applicable credit signal.

\textbf{Advantage Estimators.}
A complementary line of work analyzes the GRPO advantage estimator itself. Mroueh~\cite{mroueh2025reinforcement} derives the closed-form weighting that GRPO implicitly applies under binary rewards, $\omega^+(p)=\sqrt{(1-p)/p}$ and $\omega^-(p)=-\sqrt{p/(1-p)}$, and shows that GRPO is equivalent to an adaptive weighted contrastive loss in which rare-success rollouts receive large positive weight; this adaptive weighting is a \emph{feature} on open-answer math, where rare success corresponds to a rare reasoning trajectory. Other analyses give complementary characterizations of the estimator from cross-prompt expectation~\cite{yang2026your}, rank-bias~\cite{he2025rewarding}, KL-constrained closed-form~\cite{zhang2026gvpo}, group-size scaling~\cite{wu2025takes}, supervised-learning reframings~\cite{li2025implicit,li2026disco} angles. We adopt Mroueh's closed-form as the starting point of our analysis. 
 While the works above develop theory of the estimator within the regime where the within-group composition reflects reasoning, our analysis focuses on the complementary regime: we identify three structural cases
 in which this composition acquires a reasoning-independent component, and design an advantage variant whose magnitude is independent of this composition.

\section{Method}\label{sec:method}

Section~\ref{subsec:prelim} introduces the preliminaries of  GRPO. Section~\ref{subsec:impact} examines the learning signal within GRPO, and introduces the \emph{spurious advantage} phenomenon that has been overlooked. Section~\ref{subsec:fail-cases} diagnoses the effect of this phenomenon. Section~\ref{subsec:method} presents the proposed method.

\subsection{Preliminary}\label{subsec:prelim}

Given a question $q$ and a current policy $\pi_\theta$, GRPO~\cite{shao2024deepseekmath} samples a group of $G$ rollouts $\{\tau_1, \ldots, \tau_G\}$ from the old policy $\pi_{\theta_\text{old}}$ and assigns each rollout a binary verifier reward $r_i \in \{+1, -1\}$. We write $\npos$ and $\nneg = G - \npos$ for the in-group counts of correct and wrong rollouts, and refer to $(\npos, \nneg)$ as the within-group composition. The per-rollout advantage is computed by within-group standardization,
\begin{equation}\label{eq:grpo-adv}
\hat{A}_i = \frac{r_i - \mu}{\sigma + \varepsilon},
\end{equation}
where $\mu = \tfrac{1}{G}\sum_j r_j$, $\sigma^2 = \tfrac{1}{G}\sum_j (r_j - \mu)^2$, and $\varepsilon$ is a numerical stabilizer. The policy is then updated with a PPO-style clipped surrogate,
\begin{equation}\label{eq:grpo-loss}
\begin{aligned}
&\mathcal{L}_\text{GRPO}(\theta) =\;  -\frac{1}{G}\sum_{i=1}^{G} \frac{1}{|\tau_i|} \sum_{t=1}^{|\tau_i|} \min\bigl( w_{i,t}(\theta)\, \hat{A}_i,\\
& \mathrm{clip}(w_{i,t}(\theta), 1{-}\epsilon, 1{+}\epsilon)\, \hat{A}_i \bigr) + \beta\, \mathrm{KL}\bigl[\pi_\theta \,\|\, \pi_\text{ref}\bigr],
\end{aligned}
\end{equation}
where $w_{i,t}(\theta) = \dfrac{\pi_\theta(\tau_{i,t} \mid q, \tau_{i,<t})}{\pi_{\theta_\text{old}}(\tau_{i,t} \mid q, \tau_{i,<t})}$ is the per-token importance ratio. The advantage $\hat{A}_i$ enters Eq.~\eqref{eq:grpo-loss} as a scalar multiplier on the token-level log-likelihood ratio. Taking the gradient with respect to $\theta$ (and ignoring the KL term in the unclipped regime where $w_{i,t}{\approx}1$) gives
\begin{equation}\label{eq:grpo-grad}
\nabla_\theta \mathcal{L}_i(\theta) \;=\; -\, \hat{A}_i \cdot \frac{1}{|\tau_i|} \sum_{t=1}^{|\tau_i|} \nabla_\theta \log \pi_\theta(\tau_{i,t} \mid q, \tau_{i,<t}),
\end{equation}
where $|\hat{A}_i|$ is the per-rollout scalar weight on rollout $i$'s gradient contribution to the policy update (see Appendix~\ref{app:learning-weight} for further discussion).

\begin{figure}[t]
\centering
\includegraphics[width=0.96\columnwidth]{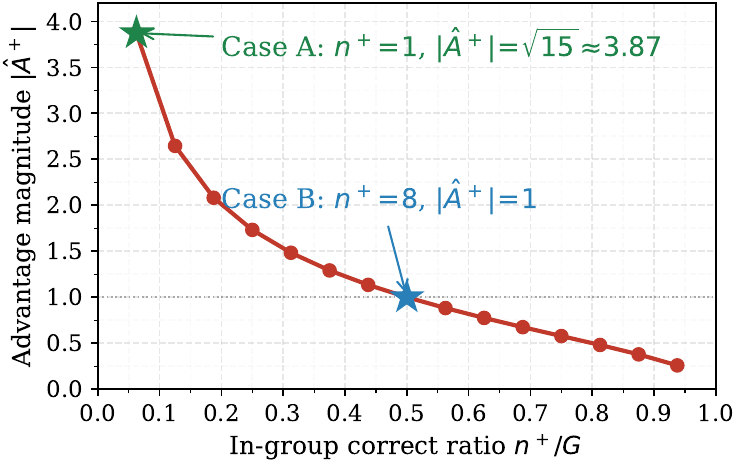}
\vspace{-1mm}
\caption{Illustration of GRPO's advantage estimator. Per-rollout magnitude $|\Apos| = \sqrt{\nneg/\npos}$ is a function of the within-group correct ratio $\npos/G$ under binary rewards. \textbf{The smaller $\npos$ is, the larger $|\Apos|$ becomes:} when a policy happens to obtain a rare correct rollout (small $\npos$) by chance rather than by reasoning, the estimator nevertheless assigns it a large $|\Apos|$, which by Eq.~\eqref{eq:grpo-grad} translates into a disproportionately high gradient weight on that rollout. We examine this effect on the learning signal in Section~\ref{subsec:impact}.}
\label{fig:motivation}
\vspace{-3mm}
\end{figure}

Substituting $r_i \in \{+1, -1\}$ into the definitions of $\mu$ and $\sigma$ and grouping by reward class, the group reward statistics can be simplified to
\begin{equation}\label{eq:mu-sigma}
\mu = \frac{\npos - \nneg}{G}, \quad \sigma = \frac{2\sqrt{\npos\,\nneg}}{G},
\end{equation}
and substituting into Eq.~\eqref{eq:grpo-adv} gives
\begin{equation}\label{eq:grpo-binary}
\Apos = \sqrt{\nneg / \npos}, \quad \Aneg = -\sqrt{\npos / \nneg},
\end{equation}
which has also been derived in concurrent work~\cite{mroueh2025reinforcement}. The magnitude assigned to any rollout is thus a function of $(\npos, \nneg)$ alone (Figure~\ref{fig:motivation}). The content of the rollout and the policy's confidence on the prompt do not enter.

\subsection{Impact of the GRPO advantage estimator}\label{subsec:impact}

The magnitude in Eq.~\eqref{eq:grpo-binary} is largest in imbalanced groups. For $G = 16$, a rare-correct group ($\npos = 1$) gives $|\Apos| = \sqrt{15} \approx 3.87$, while a balanced group ($\npos = 8$) gives $|\Apos| = 1$. The estimator amplifies the gradient on the rare-correct rollout by a factor of roughly $3.87$ over the rollout in a balanced group, even though the verifier reward is the same.

Notably, the PPO-style clip in Eq.~\eqref{eq:grpo-loss} bounds the importance ratio $w_{i,t}$ but not $|\hat A_i|$: in the common case where $w_{i,t} \approx 1$ (e.g., the first inner update before the policy drifts), the full $|\hat A_i|$ enters the gradient through Eq.~\eqref{eq:grpo-grad}. The clip therefore hardly removes the per-rollout amplification we describe, and our experiments in Section~\ref{sec:experiment} show that the amplification leads to measurable performance gaps rather than being absorbed.

To examine what learning signal this amplification carries, we decompose the per-prompt success probability $p_q$ by the source of the $+1$ outcome:
\begin{equation}\label{eq:pq-decomp}
\underbrace{p_q}_{\text{prob. of }r=+1} \;=\; \underbrace{p_{\mathrm{r}}}_{\text{from reasoning}} \;+\; \underbrace{(1 - p_{\mathrm{r}}) \cdot p_{\mathrm{g}}}_{\text{from non-reasoning}},
\end{equation}
where $p_{\mathrm{r}}$ is the probability that the policy obtains the correct answer through reasoning, and $p_{\mathrm{g}}$ is the probability of obtaining the correct answer in any other way (for example, by emitting one of a small set of candidate answers without reasoning).
The within-group count $\npos \sim \mathrm{Binomial}(G, p_q)$ then has a contribution from each source. 
 Substituting Eq.~\eqref{eq:pq-decomp} into Eq.~\eqref{eq:grpo-binary}, the magnitude $\sqrt{\nneg/\npos}$ acts on a count $\npos$ whose expectation $G p_q$ inherits both components; the rare-correct push the estimator delivers in an imbalanced group therefore consists of a learning signal driven by reasoning and a component driven by the second term of Eq.~\eqref{eq:pq-decomp}. We refer to the latter component as the \emph{spurious advantage} in the GRPO learning signal.

When $p_{\mathrm{g}} \approx 0$, the second term of Eq.~\eqref{eq:pq-decomp} vanishes and the spurious advantage is small. This is the case for open-answer mathematical reasoning~\cite{mroueh2025reinforcement,shao2024deepseekmath,guo2025deepseek} (See more in Appendix~\ref{app:open-answer}). The spurious advantage becomes sizeable when $p_{\mathrm{g}}$ is not small; the next subsection works out two such cases.

\subsection{Cases with a sizeable spurious advantage}\label{subsec:fail-cases}

\noindent\textbf{Bounded answer sets.}
When the task admits only a small set of candidate answers (e.g., multi-choice with $k$ options), a policy that emits one candidate at random hits the correct answer with probability $p_{\mathrm{g}} \approx 1/k$. For 4-choice with $G=16$, Eq.~\eqref{eq:pq-decomp} gives a baseline expectation of $G \cdot p_{\mathrm{g}} = 4$ correct rollouts from non-reasoning alone. An observed group with $\npos = 1$ sits below this baseline; yet the estimator assigns the lone correct rollout magnitude $|\Apos| = \sqrt{15} \approx 3.87$. By Eq.~\eqref{eq:pq-decomp}, the expected fraction of this magnitude is
\begin{equation}\label{eq:frac-reason-mcq}
\frac{p_{\mathrm{r}}}{p_q} \;=\; \frac{p_{\mathrm{r}}}{p_{\mathrm{r}} + (1-p_{\mathrm{r}})/k},
\end{equation}
which is small whenever $p_{\mathrm{r}}$ is comparable to or below $1/k$. The same form applies to any task with a finite candidate set, \emph{e.g.}, bounded-output classification and QA over a closed vocabulary.

A nominally open-answer training set can also exhibit the bounded case on a sizeable subset of its problems. We analyze MATH-7.5K by the syntactic shape of the gold answer, and group them into $11$ shape categories (Table~\ref{tab:math75k-category}; see Appendix~\ref{app:math75k-justification} for per-quantity derivations). Five of the $11$ categories are flagged as \emph{bounded}, together covering $55.95\%$ of MATH-7.5K, indicating that even an open-answer benchmark can contain a sizeable subset with closed answer form. Section~\ref{subsec:diagnostic} returns to this point with the per-string empirical baseline and the per-level breakdown.

\begin{table}[t]
\centering
\resizebox{\columnwidth}{!}{
\begin{tabular}{lrrc}
\hline
Category & \# / Share (\%) & Guess hit rate ($p_{\mathrm{g}}$) & Bounded \\
\hline
\texttt{int\_small} [-10,10]   & 1806 / 24.41 & 4.76\%  & \cmark \\
\texttt{int\_medium} [-100,100]& 1701 / 22.99 & 0.50\%  & \cmark \\
\texttt{expression}            & 1544 / 20.87 & 0.00\%  & --     \\
\texttt{int\_large}            & 1206 / 16.30 & 0.00\%  & --     \\
\texttt{simple\_fraction}      &  556 /  7.51 & 1.00\%  & \cmark \\
\texttt{tuple\_or\_list}       &  270 /  3.65 & 0.00\%  & --     \\
\texttt{other}                 &  121 /  1.64 & 0.00\%  & --     \\
\texttt{decimal\_short}        &  116 /  1.57 & 0.00\%  & --     \\
\texttt{finite\_set\_listed}   &   46 /  0.62 & 20.00\% & \cmark \\
\texttt{percent}               &   31 /  0.42 & 0.99\%  & \cmark \\
\texttt{empty}                 &    2 /  0.03 & 0.00\%  & --     \\
\hline
\end{tabular}}
\caption{Answer-shape distribution of MATH-7.5K. \textbf{\# / Share (\%)} reports the problem count in each shape category and its percent over the $7{,}399$ parsable problems. \textbf{Guess hit rate} is the probability that a random answer exactly matches the gold answer, i.e., $1/S$ for a category with a finite answer surface of size $S$ and $0$ when the surface is open-ended. The bounded categories cover \textbf{$55.95\%$} of MATH-7.5K. }
\vspace{-2mm}
\label{tab:math75k-category}
\end{table}

\noindent\textbf{Multi-turn search agents.}
Consider a search agent (e.g., Search-R1) that, on a prompt $q$, samples $G$ trajectories with up to $B$ search invocations per trajectory. Let $T_i \in \{1, \ldots, B\}$ denote the actual number of turns at which trajectory $i$ terminates, and let $a_i = (a_{i,1}, \ldots, a_{i, T_i})$ denote its turn-level decision sequence (when to search, what query to issue, when to answer). The reward is computed at the trajectory endpoint upon the exact-match score,
\begin{equation}\label{eq:em-reward}
r_i \;=\; \mathbf{1}\bigl[\hat{y}(a_i) = y^*\bigr], \qquad r_i \perp T_i, \; r_i \perp a_i,
\end{equation}
in the sense that $r_i$ depends on the trajectory only through its final answer $\hat{y}(a_i)$. Trajectories with very different $(T_i, a_i)$ can yield $r_i = 1$. For two trajectories $i, j$ on the same prompt with $r_i = r_j = 1$ but $(T_i, a_i) \neq (T_j, a_j)$, Eq.~\eqref{eq:grpo-binary} assigns the identical magnitude $\hat{A}_i = \hat{A}_j = \sqrt{\nneg / \npos}$. Splitting $p_q$ as in Eq.~\eqref{eq:pq-decomp}, $p_{\mathrm{g}}$ now includes the probability that the policy terminates with a correct final answer through any trajectory. This contribution can be sizeable when $B$ is not small, since the space of $(T_i, a_i)$ that can land on the correct $\hat{y}$ is large.

\subsection{Proposed method: \textsc{SignBalance}}\label{subsec:method}

We construct an advantage whose magnitude does not depend on $(\npos, \nneg)$ and which therefore does not carry the spurious component identified in Section~\ref{subsec:impact}. We refer to the resulting variant as \textsc{SignBalance} and present it in three steps. Each step removes one dependence of Eq.~\eqref{eq:grpo-binary} on $(\npos, \nneg)$. 

\paragraph{Step 1: per-class normalization.}
Normalize correct and wrong rollouts separately,
\begin{equation}\label{eq:csn}
\hat{A}_i^{(1)} \;=\; \frac{r_i - \mu^{c(i)}}{\sigma^{c(i)} + \varepsilon},
\end{equation}
where $c(i) \in \{+, -\}$ and $(\mu^{c(i)}, \sigma^{c(i)})$ are statistics inside the correct or wrong subgroup. Given such a design, a correct rollout in an imbalanced group is no longer rescaled against many wrong rollouts. However, $\sigma^{+}$ still depends on $\npos$ alone and $\sigma^{-}$ on $\nneg$ alone; the per-class magnitude remains a function of within-group counts.

\paragraph{Step 2: sign-only magnitude.}
Collapse the magnitude to a global scale,
\begin{equation}\label{eq:signonly}
\hat{A}_i^{(2)} \;=\; \mathrm{sign}(r_i) \cdot c.
\end{equation}
Such a design fully decouples the per-rollout magnitude from $(\npos, \nneg)$ and the magnitude no longer depends on within-group composition. However, the total correct-side force $\sum_{i \in +} |\hat{A}_i^{(2)}| = \npos \cdot c$ no longer matches the wrong-side force $\nneg \cdot c$ whenever $\npos \neq \nneg$, so the batch-level zero-mean property of the GRPO advantage is broken.

\paragraph{Step 3: sign with per-class force balance.}
The proposed main variant restores batch-level zero-mean without reintroducing a count-dependent per-rollout magnitude:
\begin{equation}\label{eq:level3}
\Apos = c, \qquad \Aneg = -\,c \cdot \mathrm{sg}\!\left[\tfrac{\npos}{\nneg}\right],
\end{equation}
where $\mathrm{sg}[\cdot]$ is the stop-gradient operator. The rescaling enforces $\sum_{i \in +} |\hat{A}_i^{(3)}| = \sum_{i \in -} |\hat{A}_i^{(3)}|$, while the per-rollout magnitude $c$ remains constant.
\paragraph{Summary.}
The proposed Eq.~\eqref{eq:level3} is a drop-in replacement for Eq.~\eqref{eq:grpo-adv} inside the GRPO loss Eq.~\eqref{eq:grpo-loss}. It is parameter-free apart from the global scale $c$, introduces no external model, and adds no inference cost. Since the per-rollout magnitude does not read from $(\npos, \nneg)$, the spurious advantage isolated in Section~\ref{subsec:impact} does not enter the gradient assignment.

\begin{table*}[!t]
\centering
\resizebox{\textwidth}{!}{
\begin{tabular}{l|cccc|cccc|c}
\hline
\multirow{2}{*}{Methods}
& \multicolumn{4}{c|}{Open-answer}
& \multicolumn{4}{c|}{Bounded-answer}
& \multirow{2}{*}{Avg-8} \\
\cline{2-9}
& GSM8K & MATH-500 & Min.-M & Olymp.
& MMLU-m & SAT-M & AQuA & AMC
& \\
\hline
Untrained Qwen2.5-0.5B-IT
& 47.23 & 48.40 & 8.46 & 10.25 & 52.54 & 59.38 & 31.89 & 6.02 & 33.02 \\
\hline
PPO
& 47.50 & 54.20 & 6.62 & 7.23 & 52.81 & 67.19 & 30.71 & 6.02 & 34.04 \\
REINFORCE++
& 48.83 & 55.20 & 6.99 & 7.95 & 53.61 & 64.06 & 29.92 & 7.23 & 34.22 \\
Dr.GRPO
& 48.90 & 55.60 & 6.99 & 8.13 & 52.27 & 68.75 & 25.98 & 7.23 & 34.23 \\
GRPO
& \textbf{49.89} & \textbf{56.60} & 6.25 & 7.07 & 52.94 & 65.62 & 29.53 & 6.02 & 34.24 \\
RLOO
& 49.20 & 55.80 & 7.35 & 7.60 & 52.94 & 67.19 & 30.71 & 7.23 & 34.75 \\
BNPO
& 49.51 & 56.20 & 7.35 & 8.13 & 53.61 & 68.75 & 33.46 & 8.43 & 35.68 \\
DAPO
& 48.82 & 55.40 & 6.25 & 8.48 & \textbf{55.88} & \textbf{71.88} & \textbf{36.22} & 7.23 & 36.27 \\
\hline
\rowcolor{rowhl}
\textbf{Ours}
& 49.66 & 53.60 & \textbf{8.46} & \textbf{8.83} & 54.14 & \textbf{71.88} & 35.43 & \textbf{10.84} & \textbf{36.61} \\
\hline
\end{tabular}
}
\caption{Main results on $8$ math benchmarks (Qwen2.5-0.5B-Instruct, MATH-7.5K training). Benchmarks are split into open-answer (GSM8K, MATH-500, Minerva-Math, Olympiad) and bounded-answer (MMLU-math, SAT-Math, AQuA, AMC). All methods share the same backbone, training data, and evaluation protocol; the advantage construction is the sole controlled variable. \textbf{Bold}: best per column among the RL methods (the Untrained row is a reference and is excluded from the comparison).}
\label{tab:main}
\vspace{-3mm}
\end{table*}

\section{Experiments}\label{sec:experiment}

\subsection{Experimental Settings}\label{subsec:settings}

\textbf{Backbone and training data.}
We adopt Qwen2.5-0.5B-Instruct as the primary policy and Qwen2.5-3B-Base for scale generalization, training on the MATH dataset~\cite{hendrycks2021measuring} ($\sim7{,}500$ problems) with $G = 16$ rollouts per question (max response length $8{,}192$). Optimization uses learning rate $1\times10^{-6}$, KL coefficient $\beta = 1\times10^{-3}$, up to $1000$ steps; the reward is binary, from a rule-based verifier matching the ground-truth answer.
For the search-agent setting, we adopt Qwen2.5-7B-Instruct as the policy.

\textbf{Evaluation.}
We evaluate across three settings tailored to the model capacity. \textbf{(a) 0.5B math reasoning.} We use $8$ math benchmarks, split by answer space into the \textbf{open-answer} group (free-form numeric or symbolic answer: GSM8K~\cite{cobbe2021training}, MATH-500~\cite{lightman2024let}, Minerva-Math~\cite{lewkowycz2022solving}, OlympiadBench~\cite{he2024olympiadbench}) and the \textbf{bounded-answer} group (correct answer drawn from a small set: MMLU-math~\cite{hendrycks2020measuring}, SAT-Math~\cite{zhong2024agieval}, AQuA~\cite{ling2017program}, AMC~\cite{numina_math_datasets}). We report per-benchmark accuracy and the overall mean Avg-8. \textbf{(b) 3B math reasoning.} Following~\cite{wang2025lambda}, we use a more difficult $8$-benchmark suite (GSM8K~\cite{cobbe2021training}, MATH-500~\cite{lightman2024let}, Minerva-Math~\cite{lewkowycz2022solving}, Gaokao~\cite{zhang2024mario}, OlympiadBench~\cite{he2024olympiadbench}, College Math~\cite{tang2024mathscale}, AIME24~\cite{numina_math_datasets}, AMC23~\cite{numina_math_datasets}) covering both standard and competition-grade math, and report Avg-8. \textbf{(c) Search-agent QA.} We use the $6$-benchmark text-QA following Search-R1~\cite{jin2025search}: NQ~\cite{kwiatkowski2019natural}, TriviaQA~\cite{joshi2017triviaqa}, PopQA~\cite{mallen2023not} (single-hop) and HotpotQA~\cite{yang2018hotpotqa}, 2WikiMultiHopQA~\cite{ho2020constructing}, MuSiQue~\cite{trivedi2022musique} (multi-hop). We report the mean Avg-6 upon exact match score. 

\begin{figure}[t]
\centering
\includegraphics[width=0.95\columnwidth]{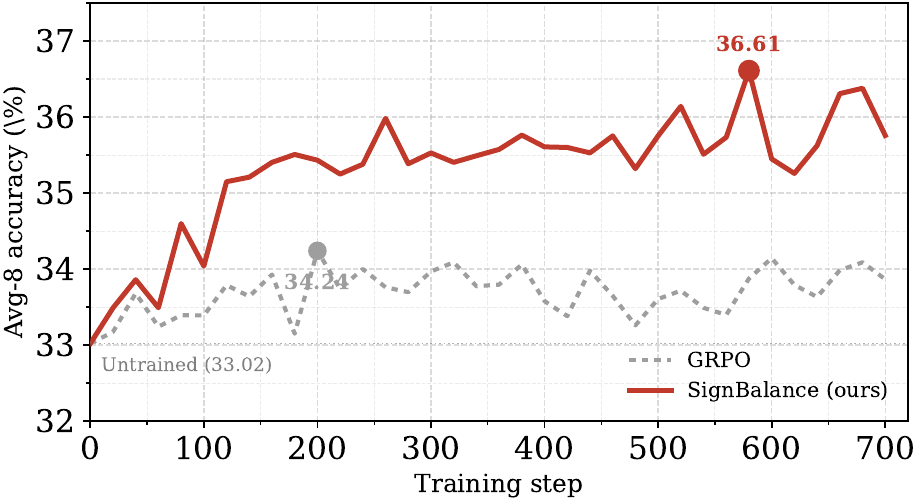}
\vspace{-1mm}
\caption{ SignBalance's Avg-8 advantage over GRPO is sustained across the entire training trajectory, not just at the best checkpoint reported in Table~\ref{tab:main}.}
\label{fig:traj_avg8}
\vspace{-3mm}
\end{figure}

\textbf{Baselines.}
The 0.5B and 3B comparisons both cover: the untrained backbone; the classical critic-based actor-critic method PPO~\cite{schulman2017proximal}; the modernized REINFORCE-style baseline REINFORCE++~\cite{hu2026reinforce++}; the group-relative baseline RLOO~\cite{ahmadian2024back}; standard GRPO~\cite{shao2024deepseekmath} and its recent variants Dr.GRPO~\cite{liu2025understanding}, DAPO~\cite{yu2026dapo}, and BNPO~\cite{xiao2025bnpo}. All methods share the same backbone, training data, optimization, and evaluation pipeline; the advantage estimation mechanism is the sole controlled variable.

\textbf{Agent setting.}
We further evaluate on Search-R1~\cite{jin2025search}, which trains a search agent with outcome-only reward (exact-match on the final answer) under a multi-turn search protocol with up to $B = 4$ search invocations per trajectory. Following Search-R1, we use Qwen2.5-7B-Instruct as the policy, adopt the same training corpus, and report exact-match accuracy at evaluation.
See more implementation details, in  Appendix~\ref{app:implementation}.

\begin{table*}[t]
\centering
\resizebox{\textwidth}{!}{
\begin{tabular}{l|ccccccccc}
\hline
Methods & GSM8K & MATH-500 & Min.-M & Gaokao & Olymp. & C.\,Math & AIME & AMC & Avg-8 \\
\hline
Untrained Qwen2.5-3B-Base
& 81.6 & 58.8 & 25.0 & 47.0 & 24.1 & 32.8 & 6.5 & 35.8 & 38.90 \\
\hline
PPO
& 84.8 & 62.5 & 30.0 & 51.4 & 26.5 & 36.0 & 6.2 & 35.3 & 41.59 \\
REINFORCE++
& 85.3 & 63.5 & 30.6 & 52.4 & 27.2 & 36.5 & 7.2 & 35.7 & 42.30 \\
Dr.GRPO
& 85.5 & 63.6 & 30.8 & 52.6 & 27.4 & 36.7 & 7.5 & 36.0 & 42.51 \\
DAPO
& 85.1 & \textbf{65.4} & 31.2 & 53.0 & 27.6 & 36.6 & 6.2 & 35.9 & 42.60 \\
GRPO
& 85.7 & 64.0 & 31.2 & 53.0 & 27.9 & 37.0 & 7.7 & 36.5 & 42.80 \\
RLOO
& 85.9 & 64.3 & 31.4 & 53.2 & 28.0 & 37.1 & 7.8 & 36.7 & 43.05 \\
BNPO
& 86.0 & 64.4 & 31.4 & 53.4 & 28.0 & 37.2 & 8.0 & 37.0 & 43.18 \\
\hline
\rowcolor{rowhl}
\textbf{Ours}
& \textbf{86.4} & 64.8 & \textbf{32.0} & \textbf{54.2} & \textbf{28.6} & \textbf{37.5} & \textbf{8.5} & \textbf{38.2} & \textbf{43.78} \\
\hline
\end{tabular}}
\caption{Scale generalization on Qwen2.5-3B-Base across an 8-benchmark math suite tailored to the 3B scale (following~\cite{wang2025lambda}); see Section~\ref{subsec:settings} for the bench list and the rationale for the bench-set change relative to Table~\ref{tab:main}. \textbf{Bold}: best per column among the RL methods (the Untrained row is a reference).}
\label{tab:scale-3b}
\vspace{-2mm}
\end{table*}

\begin{table*}[t]
\centering
\resizebox{0.95\textwidth}{!}{
\begin{tabular}{l|ccc|ccc|c}
\hline
\multirow{2}{*}{Methods}
& \multicolumn{3}{c|}{Single-hop}
& \multicolumn{3}{c|}{Multi-hop}
& \multirow{2}{*}{Avg-6} \\
\cline{2-7}
& NQ & TriviaQA & PopQA
& HotpotQA & 2Wiki & MuSiQue
& \\
\hline
R-Search~\cite{zhao2025r}
& 34.71 & 58.85 & 36.17 & 32.59 & 29.76 & 12.83 & 34.15 \\
ZeroSearch~\cite{sun2025zerosearch}
& 36.29 & 57.09 & 35.79 & 32.11 & 34.90 & 10.88 & 34.51 \\
DeepResearcher~\cite{zheng2025deepresearcher}
& 35.60 & 58.44 & 37.06 & 34.85 & 31.39 & 14.40 & 35.29 \\
ReSearch~\cite{chen2025learning}
& 36.40 & 58.71 & 38.87 & 34.88 & 27.97 & 17.09 & 35.65 \\
Search-R1~\cite{jin2025search}
& 39.20 & 59.22 & 38.91 & 36.16 & 27.58 & 14.94 & 36.00 \\
StepSearch~\cite{wang2025stepsearch}
& 36.84 & 57.51 & 36.41 & 36.31 & 32.05 & \textbf{19.53} & 36.44 \\
\hline
\rowcolor{rowhl}
\textbf{Ours}
& \textbf{39.25} & \textbf{59.30} & \textbf{39.00} & \textbf{36.55} & \textbf{35.20} & 17.50 & \textbf{37.80} \\
\hline
\end{tabular}
}
\caption{Exact-match accuracy on six text-based QA benchmarks under the search-agent scenario (Qwen2.5-7B-Instruct, Section~\ref{subsec:settings}). All methods use E5-base-v2 with Wikipedia-18 as the retriever and identical inference parameters; the advantage estimation mechanism within the GRPO outer loop is the sole controlled variable.}
\label{tab:search_agent}
\vspace{-3mm}
\end{table*}

\subsection{Main Results}\label{subsec:main-results}

Table~\ref{tab:main} reports the per-benchmark accuracy of GRPO, the GRPO variants, and \textsc{SignBalance}. We observe two main patterns.
On bounded-answer benchmarks, \textsc{SignBalance} attains the large improvements, \emph{e.g.}, SAT-Math $71.88$ vs.\ GRPO $65.62$ ($+6.26$) and  AQuA $35.43$ vs.\ $29.53$ ($+5.90$). A plausible explanation is that when the within-group composition contains a sizeable random-guess component, the count-dependent magnitude in Eq.~\eqref{eq:grpo-binary} amplifies that component, while the constant magnitude of \textsc{SignBalance} does not. Appendix~\ref{app:trace} provides side-by-side reasoning traces in which the GRPO-trained policy produces a correct derivation but selects a wrong letter, while the \textsc{SignBalance}-trained policy maps the same derivation to the correct letter. Second, on open-answer benchmarks, \textsc{SignBalance} brings clear improvements over the untrained baseline.

\begin{figure}[t]
\centering
\includegraphics[width=0.98\columnwidth]{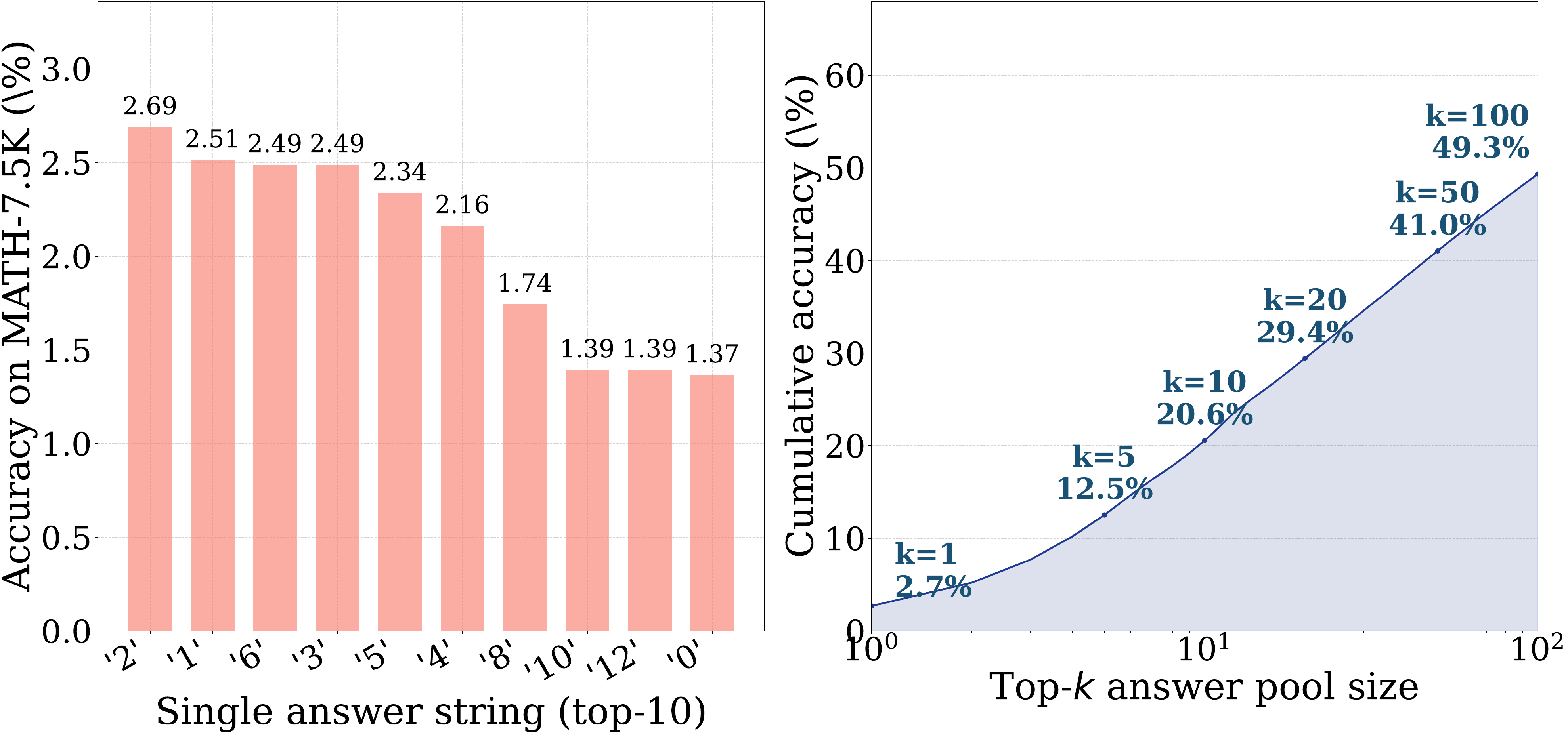}
\vspace{-1mm}
\caption{A policy that does no reasoning at all can still score on MATH-7.5K just by guessing popular answer strings. \textbf{Left}: per-string accuracy of a constant policy that always emits one fixed string from the top-$10$ list (emitting ``$2$'' alone scores $2.69\%$). \textbf{Right}: cumulative accuracy under uniform sampling over the top-$k$ strings ($20.6\%$ at $k{=}10$, ${\sim}50\%$ at $k{=}100$). Even on a free-form math benchmark, there is a measurable lower-bound score from guessing, amplifying the gradient-scalar weight defined in Eq.~\eqref{eq:grpo-grad}.}
\label{fig:math75k-modal}
\vspace{-3mm}
\end{figure}

\begin{figure}[t]
\centering
\includegraphics[width=0.98\columnwidth]{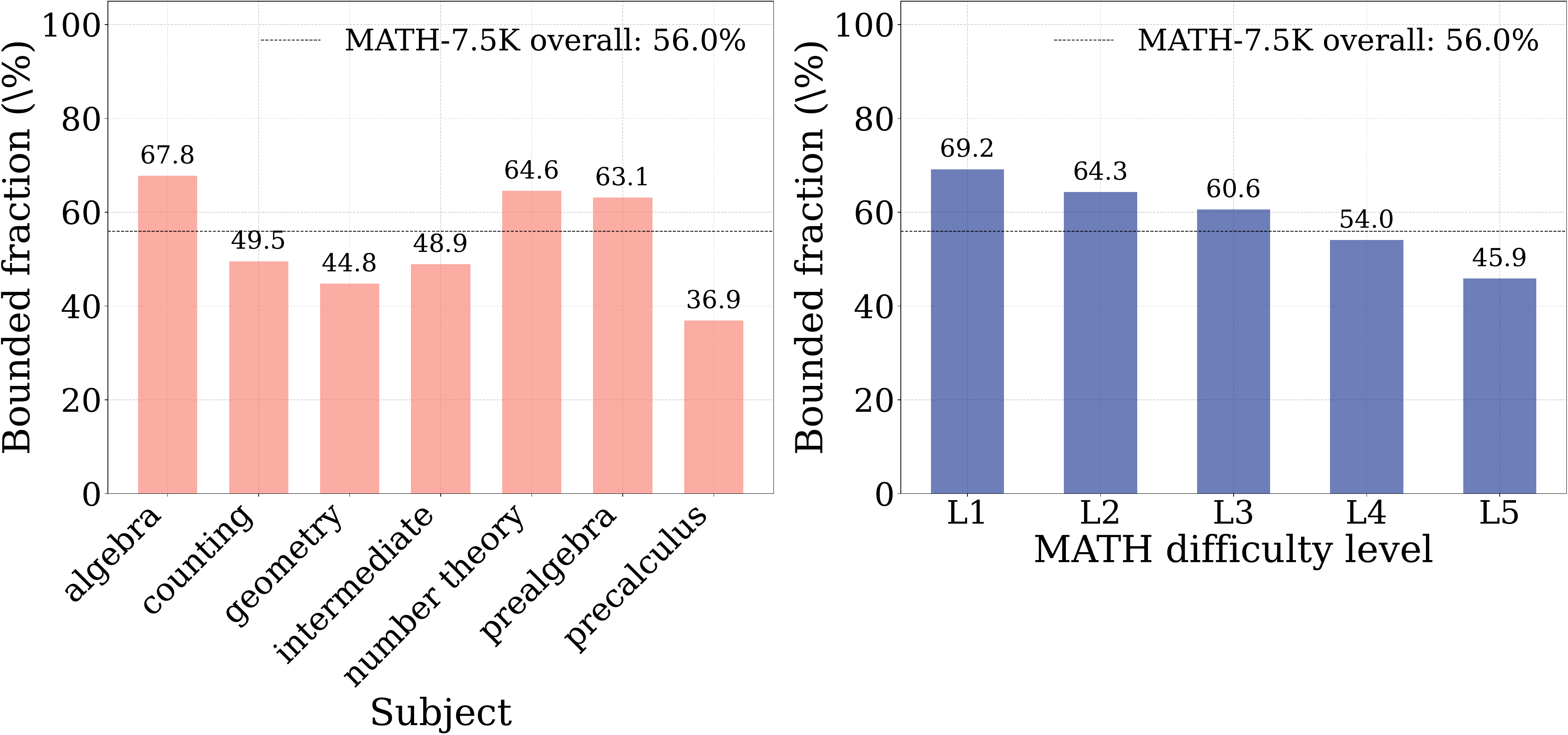}
\vspace{-3mm}
\caption{The bounded-answer fraction concentrates on the easy subset of MATH-7.5K. \textbf{Left}: bounded-answer fraction by subject. \textbf{Right}: bounded-answer fraction by MATH difficulty level. The fraction drops monotonically from $69.2\%$ at Level~1 to $45.9\%$ at Level~5; easier problems tend to have small answer surfaces (e.g., $0$--$10$ integers), which are exactly the problems an early-stage policy is most likely to solve, amplifying the spurious advantage during early training.}
\label{fig:math75k-level}
\vspace{-2mm}
\end{figure}

\begin{figure*}[t]
\centering
\includegraphics[width=0.95\textwidth]{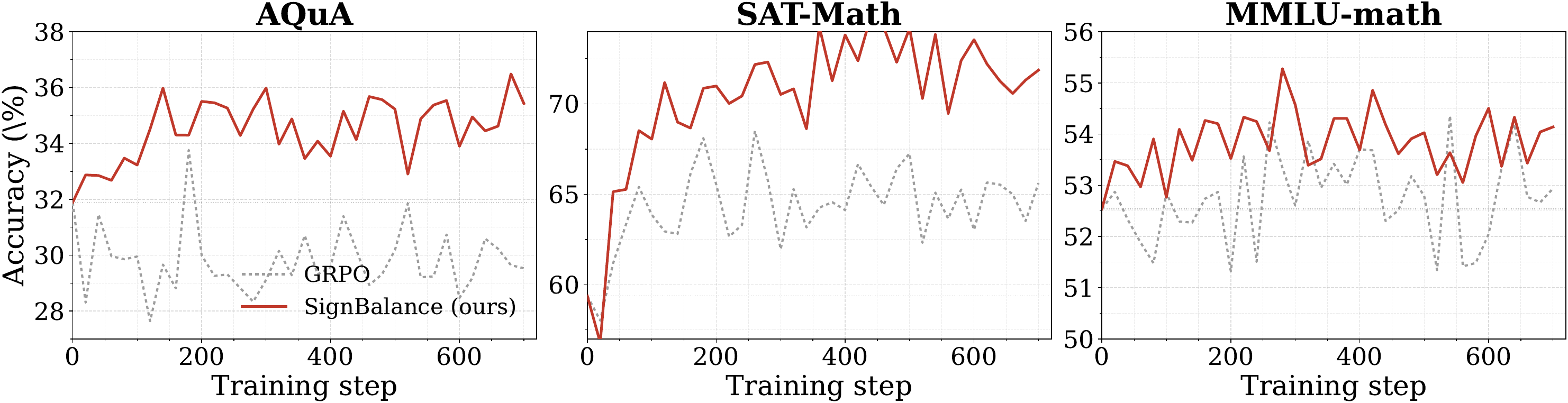}
\caption{\textbf{The per-bench gains of \textsc{SignBalance} over GRPO on bounded-answer benchmarks are sustained across the training trajectory, not concentrated at a single checkpoint.} Per-checkpoint accuracy on AQuA, SAT-Math, and MMLU-math; best-checkpoint values match the corresponding columns of Table~\ref{tab:main}.}
\label{fig:traj_bounded}
\vspace{-2mm}
\end{figure*}

\begin{table*}[t]
\centering
\resizebox{\textwidth}{!}{
\begin{tabular}{l|cccc|cccc|c}
\hline
\multirow{2}{*}{Variants}
& \multicolumn{4}{c|}{Open-answer}
& \multicolumn{4}{c|}{Bounded-answer}
& \multirow{2}{*}{Avg-8} \\
\cline{2-9}
& GSM8K & MATH-500 & Min.-M & Olymp.
& MMLU-m & SAT-M & AQuA & AMC
& \\
\hline
GRPO baseline
& 49.89 & 56.60 & 6.25 & 7.07 & 52.94 & 65.62 & 29.53 & 6.02 & 34.24 \\
\hline
Sep.\ norm (Step 1)
& 48.50 & \textbf{55.80} & 6.80 & 7.60 & 53.61 & 62.50 & 31.10 & 9.20 & 34.39 \\
Asym.\ boost
& 46.93 & 53.40 & 6.99 & 7.95 & 52.94 & 71.88 & 31.89 & 8.43 & 35.05 \\
Sign-only (Step 2)
& 47.99 & 51.80 & 7.72 & 8.13 & 53.61 & 71.88 & 29.92 & 9.64 & 35.09 \\
\hline
\rowcolor{rowhl}
\textbf{Ours (Step 3)}
& \textbf{49.66} & 53.60 & \textbf{8.46} & \textbf{8.83} & \textbf{54.14} & \textbf{71.88} & \textbf{35.43} & \textbf{10.84} & \textbf{36.61} \\
\hline
\end{tabular}}
\caption{Ablation on the advantage estimation. ``Sep.\ norm'' is Step~1 of \textsc{SignBalance} (class-separate normalization, Section~\ref{subsec:method}); ``Sign-only'' is Step~2 ($\Apos = \Aneg = c$, no force balance); ``Asym.\ boost'' is a fixed-factor variant ($\Apos = 2.5$, $\Aneg = -1$) included to isolate the role of force balance: it also removes the per-rollout count dependence, but with a hand-coded asymmetry that breaks the batch-level zero-mean property; ``Ours'' is the full \textsc{SignBalance} (Step~3, sign + per-class force balance).
}
\label{tab:ablation}
\vspace{-3mm}
\end{table*}

\textbf{Scale generalization to 3B.}
We reproduce the same setup on Qwen2.5-3B-Base using the benchmarks introduced in Section~\ref{subsec:settings}. As shown in Table~\ref{tab:scale-3b}, \textsc{SignBalance} again attains the highest Avg-8 ($43.78$), with the largest per-bench margins on the competition and bounded-answer cases (Gaokao $+1.2$, AMC $+1.7$, AIME $+0.8$). The per-bench leader count grows from $4/8$ at $0.5$B to $7/8$ at $3$B, so the advantage of \textsc{SignBalance} broadens across benchmarks at the larger scale rather than washing out, even as the absolute Avg-8 gap over GRPO narrows from $+2.37$ to $+0.98$.

\subsection{Empirical Analysis and Discussion}\label{subsec:diagnostic}

We analyze where the spurious advantage component is large and how it affects the GRPO estimator's training behavior, on the 0.5B setting.\footnote{We further quantify three additional diagnostics --- the random-guess share of correct rollouts, the checkpoint-to-checkpoint stability of GRPO's evaluation, and the performance gain split by answer-space regime --- in Appendix~\ref{app:additional-diagnostics}.}

Figure~\ref{fig:math75k-modal} shows that a policy that emits no reasoning at all can score on MATH-7.5K by guessing popular answers: constantly outputting the single string ``$2$'' scores $2.69\%$, and uniformly sampling from the top-$10$ strings reaches $20.6\%$. Figure~\ref{fig:math75k-level} shows the bounded-answer fraction is highest on easy problems (Level~1: $69.2\%$ $\to$ Level~5: $45.9\%$), so the spurious advantage component is largest on exactly the problems an early-stage policy is most likely to solve. Together with Table~\ref{tab:math75k-category} in Section~\ref{subsec:fail-cases}, these observations confirm that the bounded-answer case is not specific to MCQ benchmarks: it shows up inside what is widely treated as a free-form math training set.

\subsection{Multi-Turn Agent Training}\label{subsec:scope-agent}

Table~\ref{tab:search_agent} reports exact-match accuracy under the search-agent scenario described in Section~\ref{subsec:settings}, where the multi-turn trajectory structure of Section~\ref{subsec:fail-cases} applies. We compare \textsc{SignBalance} against six representative search-agent methods: Search-R1~\cite{jin2025search}, StepSearch~\cite{wang2025stepsearch}, ZeroSearch~\cite{sun2025zerosearch}, R-Search~\cite{zhao2025r}, ReSearch~\cite{chen2025learning}, and DeepResearcher~\cite{zheng2025deepresearcher}.

\textsc{SignBalance} attains the highest Avg-6 ($37.80$, $+1.36$ over StepSearch and $+1.80$ over Search-R1) and is the per-bench leader on $5$ of the $6$ benchmarks. The largest single-bench gain over Search-R1 appears on 2WikiMultiHopQA ($35.20$ vs.\ $27.58$, $+7.62$), the most multi-hop-intensive bench in the suite. The one bench where \textsc{SignBalance} is not the per-bench leader is MuSiQue, where StepSearch's step-level reward shaping (a procedure-level addition orthogonal to advantage-estimator design) holds the top spot.
The advantage estimator in vanilla GRPO can encourage the policy to guess the correct answer, fitting the underlying statistics and lacks generalization ability. The proposed method enhances the generalization, benefiting the broader scenarios, when there are same outer GRPO loop, same training corpus, and only the per-rollout advantage differs.

\subsection{Ablation on the Advantage Estimation}\label{subsec:ablation}

We compare the three steps of \textsc{SignBalance} (Section~\ref{subsec:method}) against the GRPO baseline and against a fixed asymmetric boost following Table~\ref{tab:main}.

Step~1 improves AQuA a bit  and brings a marginal gain over GRPO, contributing to the cross-class scale interaction. Step~2 removes the per-rollout count dependence and improves SAT-Math by $+6.26$ over GRPO and AMC by $+3.62$, but reduces MATH-500 by $4.80$ due to the loss of batch-level force balance. This matches the intuition in Section~\ref{subsec:fail-cases} where the spurious advantage peaks on bounded-answer tasks. The asymmetric boost is an additional factor that also removes the per-rollout count dependence by replacing the matched force balance with a hand-coded asymmetry. We hypothesis that the $1.56$ gap between Asym.\ ($35.05$) and Ours ($36.61$) could be caused by the hand-coded asymmetry, indicating  that force balance should be data-adaptive but not fixed. \textsc{SignBalance} (Step~3) combines sign with per-class force balance, achieving more gains on the bounded-answer benchmarks (AQuA, AMC). The averaged gain can stems from bounded-answer benchmarks. Besides, MATH-500 is the only trade-off due to the negligible effect of  the spurious advantage.

\subsection{Stability Analysis}\label{subsec:traj-analysis}

Figure~\ref{fig:traj_bounded} shows the per-checkpoint accuracy of GRPO and \textsc{SignBalance} on the three bounded-answer benchmarks (AQuA, SAT-Math, MMLU-math). The two trajectories separate early (within ${\sim}100$ steps) and \textsc{SignBalance} stays above GRPO for the rest of training on all three. This rules out the alternative reading that the per-bench gaps reported in Table~\ref{tab:main} come from a single lucky checkpoint: the gap is sustained across hundreds of training steps, supporting the picture in Section~\ref{subsec:fail-cases} that the per-rollout count dependence of GRPO consistently injects a spurious component on bounded-answer benchmarks, which the proposed method \textsc{SignBalance} removes.

\section{Conclusion}\label{sec:conclusion}

We identified a \emph{spurious advantage} component in the GRPO advantage estimator: whenever a task admits a non-zero probability of producing the correct answer without reasoning, GRPO's count-dependent magnitude amplifies that component. We worked out the two cases where it becomes sizeable (bounded answer sets and multi-turn search agents) and proposed \textsc{SignBalance}, a one-line advantage variant that decouples the per-rollout magnitude from within-group composition. \textsc{SignBalance} matches GRPO on open-answer math, beats it on bounded-answer math at both $0.5$B and $3$B, and gives the best Avg on multi-turn QA.

\section*{Limitations}
Although we have validated \textsc{SignBalance} on math reasoning and the multi-turn search-agent setting, the validation does not yet extend to more general settings. For instance, when an agent must select the most suitable tool from a finite tool library at every action, the setting closely resembles the multi-choice case in math reasoning; by our analysis, canonical GRPO may likewise suffer from the spurious advantage there, and our method is therefore expected to bring consistent gains in such a setting as well.

While we have validated \textsc{SignBalance} across settings (math reasoning and search-agent) and across model scales (0.5B, 3B, and 7B), and we empirically observe that our method drives the policy toward more effective reasoning strategies, we have not yet quantitatively measured the resulting growth in reasoning ability, such as the policy's reasoning confidence and its uncertainty behavior across problem difficulty levels.

\bibliography{main}

\clearpage
\appendix

\section{Open-answer mathematical reasoning: the small-spurious-advantage case}\label{app:open-answer}

We work out the case $p_{\mathrm{g}} \approx 0$ deferred from Section~\ref{subsec:impact}.

\paragraph{Setup.}
Consider a prompt $q$ on which the policy attains per-sample success probability $p_q$, and recall the decomposition of Eq.~\eqref{eq:pq-decomp},
\[
p_q \;=\; p_{\mathrm{r}} \;+\; (1 - p_{\mathrm{r}}) \cdot p_{\mathrm{g}},
\]
where $p_{\mathrm{r}}$ is the probability that the policy obtains the correct answer through reasoning, and $p_{\mathrm{g}}$ is the probability of obtaining the correct answer by any other route. For open-answer mathematical reasoning, the answer surface is large (numbers, fractions, symbolic expressions, tuples, intervals), and the prompt typically admits a single correct surface form; producing this form without reasoning is a $\ll 1$ event, so $p_{\mathrm{g}} \to 0$.

\paragraph{Effect on the within-group count.}
With $p_{\mathrm{g}} \to 0$, the per-prompt success probability reduces to $p_q \approx p_{\mathrm{r}}$, and the within-group count $\npos \sim \mathrm{Binomial}(G, p_q)$ has $\mathbb{E}[\npos] = G p_q \approx G p_{\mathrm{r}}$ — the count is driven by the reasoning channel alone. Substituting into Eq.~\eqref{eq:grpo-binary}, the rare-correct magnitude
\[
|\Apos| \;=\; \sqrt{\nneg / \npos}
\]
is large precisely when $p_{\mathrm{r}}$ is small, i.e., when the policy is reasoning through the prompt only rarely.

\paragraph{Connection to prior work.}
This is the case in which the closed-form weighting of Mroueh~\cite{mroueh2025reinforcement},
\[
\omega^+(p) = \sqrt{(1-p)/p}, \qquad \omega^-(p) = -\sqrt{p/(1-p)},
\]
admits the adaptive contrastive interpretation: a rare-correct rollout receives a large weight because reasoning success on this prompt is genuinely rare. When $p_{\mathrm{g}} = 0$, the second term of Eq.~\eqref{eq:pq-decomp} contributes nothing, and the magnitude assigned by Eq.~\eqref{eq:grpo-binary} corresponds to a learning signal with no spurious advantage.

\paragraph{Bound on the residual spurious advantage.}
When $p_{\mathrm{g}}$ is small but not zero, the expected fraction of the magnitude attributable to reasoning is
\[
\frac{p_{\mathrm{r}}}{p_q} \;=\; \frac{p_{\mathrm{r}}}{p_{\mathrm{r}} + (1-p_{\mathrm{r}}) p_{\mathrm{g}}},
\]
which is $\geq 1 - p_{\mathrm{g}}/p_q$. For open-answer mathematical reasoning the ratio is close to $1$, and the spurious advantage is at the level of $p_{\mathrm{g}}$ at most. The within-MATH-7.5K analysis in Section~\ref{subsec:diagnostic} quantifies this residual: even on a nominally open-answer training set, $p_{\mathrm{g}}$ is not zero and the residual is measurable, though small relative to the bounded-answer cases of Section~\ref{subsec:fail-cases}.

\section{$|\hat A_i|$ as a per-rollout learning weight}\label{app:learning-weight}

Throughout the main paper we describe $|\hat A_i|$ as the per-rollout scalar weight on rollout $i$'s gradient contribution to the policy update. Strictly, the GRPO policy gradient for rollout $i$ (ignoring the KL term and in the unclipped regime where $w_{i,t} \approx 1$) factorizes as
\[
\nabla_\theta \mathcal{L}_i \;=\; -\, \hat A_i \cdot \underbrace{\frac{1}{|\tau_i|} \sum_t \nabla_\theta \log \pi_\theta(\tau_{i,t})}_{\text{rollout-specific log-prob gradient}},
\]
so the gradient norm contributed by rollout $i$ factorizes as $|\hat A_i| \cdot \bigl\|\tfrac{1}{|\tau_i|}\sum_t \nabla_\theta \log \pi_\theta(\tau_{i,t})\bigr\|$. The second factor depends on the rollout's tokens and the current policy, but is independent of the choice of advantage estimator. For a fixed rollout, $|\hat A_i|$ therefore acts as a multiplicative scalar that determines how strongly that rollout's log-probability is pushed.

This usage matches the standard convention in the RL-for-LLM literature, where $|\hat A_i|$ is treated as the per-rollout learning weight when comparing advantage estimators~\cite{schulman2017proximal,ahmadian2024back,hu2026reinforce++,shao2024deepseekmath,mroueh2025reinforcement}. In particular, Mroueh~\cite{mroueh2025reinforcement} analyzes GRPO at this per-rollout weight level under binary rewards and derives the same $\sqrt{n^-/n^+}$ form we use in Eq.~\eqref{eq:grpo-binary}. Our analysis of the spurious advantage component is therefore a statement about how this per-rollout weight is composed across reasoning-driven and non-reasoning success channels, not a claim about gradient norms in absolute units or about the precise effect under specific PPO clip realizations.

\section{Justification of the MATH-7.5K answer-shape categorization}\label{app:math75k-justification}

This appendix walks through every quantity in Table~\ref{tab:math75k-category} column by column, so each cell can be re-derived from the underlying definitions.

\paragraph{Source set (the denominator of every column).}
We work with the standard MATH training split of $7{,}500$ problems. We extract the gold answer of each problem by parsing the \texttt{$\backslash$boxed\{...\}} token in the official solution. $7{,}399$ problems contain a parsable \texttt{$\backslash$boxed\{\}}; the remaining $101$ are dropped where parsing fails. Every column of Table~\ref{tab:math75k-category} is computed over this set of $7{,}399$ problems.

\paragraph{Category (the row label).}
Each parsable gold answer is assigned to one of $11$ surface-syntactic categories, defined purely by the shape of the answer string rather than the underlying problem domain. For example, an answer of $5$ goes to \texttt{int\_small} regardless of whether the problem is about geometry or arithmetic; an answer of $p/q$ with $|p|,|q|\leq 20$ goes to \texttt{simple\_fraction}; a non-numeric algebraic answer such as $x^2{+}1$ goes to \texttt{expression}. The $11$ categories are mutually exclusive and exhaustive, with \texttt{other} as a catch-all.

\paragraph{\#Problems (column 2).}
The count of problems in the source set whose gold answer falls in this category. The $11$ counts sum to $7{,}399$.

\paragraph{Share (column 3).}
\#Problems divided by $7{,}399$, expressed as a percent. The $11$ shares sum to $100\%$. This column tells the reader how large each shape category is relative to the whole training set.

\paragraph{Guess hit rate (column 4).}
The probability that a policy producing a uniformly random answer of the given syntactic shape lands on the gold answer. Two cases:
\begin{itemize}[nosep,leftmargin=1.5em]
    \item \emph{Finite surface.} If the category has a finite answer surface of size $S$, the hit rate is $1/S$. For \texttt{int\_small} ($S{=}21$ integers in $[-10,10]$), this gives $1/21 \approx 4.76\%$; for \texttt{int\_medium} ($S{=}201$), $1/201 \approx 0.50\%$; for \texttt{simple\_fraction} ($S{\approx}100$ irreducible fractions with $|p|,|q|\leq 20$), $\approx 1.00\%$; for \texttt{finite\_set\_listed} (typical $S{=}5$ enumerated options), $\approx 20.00\%$; for \texttt{percent} ($S{=}101$ percent values), $\approx 0.99\%$.
    \item \emph{Open-ended surface.} If the category has an infinite or open-ended answer surface (e.g., \texttt{expression}, \texttt{int\_large}, \texttt{tuple\_or\_list}, \texttt{decimal\_short}, \texttt{other}, \texttt{empty}), the hit rate is treated as $0$ — emitting a random expression or large integer that exactly matches the gold answer has negligible probability.
\end{itemize}
This column is the per-category estimate of $p_{\mathrm{g}}$ in Eq.~\eqref{eq:pq-decomp}: the probability of obtaining the correct answer without reasoning.

\paragraph{Bounded (column 5).}
A binary flag set to $\cmark$ when the guess hit rate is $\geq 0.5\%$, and $-$ otherwise. The threshold is conservative: we only flag categories where a no-reasoning policy already has a measurable hit rate, not categories with a marginal rate that would be hard to distinguish from sampling noise. Under this rule, $5$ of the $11$ categories are flagged (\texttt{int\_small}, \texttt{int\_medium}, \texttt{simple\_fraction}, \texttt{finite\_set\_listed}, \texttt{percent}), together covering $4{,}140$ problems ($55.95\%$ of MATH-7.5K).

\paragraph{Why this analysis is defensible.}
Three points clarify the reading of Table~\ref{tab:math75k-category}: (i) the categorization is syntactic, not semantic --- we group answers by surface shape, not by problem difficulty, so the table reports a fact about the answer surface of MATH-7.5K rather than a re-interpretation of the benchmark; (ii) the $0.5\%$ threshold flags only categories with a measurable hit rate, and the largest bounded category, \texttt{int\_small} ($24.41\%$ of the data), has a guess rate of $4.76\%$, well above the threshold; (iii) we do not claim that bounded problems in MATH-7.5K are easy or unimportant --- we claim only that the bounded case of Section~\ref{subsec:fail-cases} arises on a sizeable subset of MATH-7.5K, so the spurious advantage component is non-zero even on a benchmark conventionally treated as open-answer. Section~\ref{subsec:diagnostic} provides further empirical evidence.

\section{Implementation details}\label{app:implementation}

\textsc{SignBalance} sets $\Apos = 1$ and $\Aneg = -\mathrm{sg}[\npos/\nneg]$ on top of the standard GRPO policy-gradient loss, with no additional hyperparameter beyond the global scale $c = 1$. We adopt vLLM for rollout generation, and report the best checkpoint per run by Avg-8.

\section{Additional empirical diagnostics}\label{app:additional-diagnostics}

We report three additional diagnostics on the $0.5$B evaluation suite that further support the analysis in Section~\ref{subsec:diagnostic}.

\paragraph{Random-guess share of correct rollouts.}
On each benchmark, we quantify how much of a correct rollout could already be produced by a random-guess policy. For a $k$-option multi-choice benchmark, a uniform random-guess policy attains accuracy $\geq 1/k$. Under a first-order uniform-guess assumption, the random-guess share of correct rollouts is $\tfrac{(1/k)(1-s)}{\mathrm{acc}}$, where $s=(\mathrm{acc}-1/k)/(1-1/k)$. On the untrained Qwen2.5-0.5B-Instruct, this share is $53.4\%$ on AQuA ($k{=}5$), $30.1\%$ on MMLU-math ($k{=}4$), and $22.8\%$ on SAT-Math ($k{=}4$); on the open-answer math benchmarks (MATH-500, GSM8K, Minerva-Math, Olympiad), the random-guess floor is $0$ and the share is $0$. The bounded-answer benchmarks therefore feed the estimator a within-group composition that is partially random-guess even at the start of training.

\paragraph{Checkpoint-to-checkpoint stability of GRPO evaluation.}
For a GRPO training run, we compute the standard deviation of evaluation accuracy across the last $30$ training checkpoints on each benchmark (the same checkpoints used for best-checkpoint selection in Table~\ref{tab:main}). The mean standard deviation is $2.69$ on the bounded-answer subset (SAT-Math, MMLU-math, AQuA) and $0.92$ on the open-answer subset (MATH-500, GSM8K, Minerva-Math, Olympiad), a ratio of $2.91\times$. On the same training run, the GRPO estimator's late-stage evaluation moves about three times more from checkpoint to checkpoint on bounded-answer benchmarks than on open-answer ones, consistent with the spurious component contributing a noisier per-update gradient on the bounded-answer regime.

\paragraph{Performance gain by regime.}
We measure GRPO's performance gain across the two regimes via $(\mathrm{acc}_{\mathrm{late}}-1/k)-(\mathrm{acc}_{\mathrm{untr}}-1/k)$ for bounded-answer benchmarks (or simply $\mathrm{acc}_{\mathrm{late}}-\mathrm{acc}_{\mathrm{untr}}$ for open-answer ones), where $\mathrm{acc}_{\mathrm{late}}$ is the mean over the last $30$ training checkpoints. On the two open-answer benchmarks where the $0.5$B policy has enough capacity to actually improve, GRPO's gain is $+7.17$ on MATH-500 and $+2.70$ on GSM8K; on the bounded-answer benchmarks the gains are $-0.24$ on MMLU-math, $+2.26$ on AQuA, and $+5.28$ on SAT-Math. The bounded-answer gains are visibly smaller and noisier than the open-answer ones, consistent with the spurious component diluting the GRPO update on the bounded-answer regime.

\section{Algorithmic comparison with GRPO}\label{app:algo}

We provide side-by-side pseudocode of the standard GRPO advantage estimator (Algorithm~\ref{alg:grpo}) and \textsc{SignBalance} (Algorithm~\ref{alg:signbalance}). Lines in \textcolor{red}{red} mark where \textsc{SignBalance} differs from GRPO, so the reader can locate the design at a glance.

\begin{algorithm}[t]
\caption{GRPO advantage estimator (standard, per prompt $q$).}
\label{alg:grpo}
\begin{algorithmic}[1]
\Require rollouts $\{\tau_i\}_{i=1}^{G}$, binary rewards $\{r_i\}_{i=1}^{G}$, $r_i \in \{+1,-1\}$
\State $\mu \gets \tfrac{1}{G}\sum_{i} r_i$
\State $\sigma \gets \sqrt{\tfrac{1}{G}\sum_{i}(r_i - \mu)^2}$
\For{$i = 1, \ldots, G$}
    \State $\hat{A}_i \gets (r_i - \mu) / (\sigma + \varepsilon)$ \Comment{per-rollout magnitude depends on $(\npos, \nneg)$}
\EndFor
\State \Return $\{\hat{A}_i\}_{i=1}^{G}$
\end{algorithmic}
\end{algorithm}

\begin{algorithm}[t]
\caption{\textsc{SignBalance} advantage estimator (ours, per prompt $q$).}
\label{alg:signbalance}
\begin{algorithmic}[1]
\Require rollouts $\{\tau_i\}_{i=1}^{G}$, binary rewards $\{r_i\}_{i=1}^{G}$, $r_i \in \{+1,-1\}$, global scale $c$
\State \textcolor{red}{$\npos \gets |\{i : r_i = +1\}|, \quad \nneg \gets G - \npos$}
\State \textcolor{red}{$\rho \gets \mathrm{sg}[\,\npos / \nneg\,]$} \Comment{stop-gradient per-class force-balance ratio}
\For{$i = 1, \ldots, G$}
    \If{$r_i = +1$}
        \State \textcolor{red}{$\hat{A}_i \gets c$} \Comment{constant magnitude, free of $(\npos, \nneg)$}
    \Else
        \State \textcolor{red}{$\hat{A}_i \gets -\,c \cdot \rho$} \Comment{per-class rescaling restores zero-mean balance}
    \EndIf
\EndFor
\State \Return $\{\hat{A}_i\}_{i=1}^{G}$
\end{algorithmic}
\end{algorithm}

\section{Reasoning trace comparisons on bounded-answer benchmarks}\label{app:trace}

To illustrate how \textsc{SignBalance} changes the policy's behavior on bounded-answer tasks, we compare reasoning traces from the GRPO-trained and \textsc{SignBalance}-trained Qwen2.5-0.5B-Instruct policies on two test problems where the two methods disagree: one from MMLU-math, and one from SAT-Math. In both cases, the GRPO-trained policy carries out a correct derivation but then maps it to the wrong letter --- a failure that we interpret as the policy still relying on a spurious cue at the letter-selection step. The \textsc{SignBalance}-trained policy produces a closely matched derivation but maps it to the correct letter.

\paragraph{Example 1 (MMLU-math).}
\emph{Prompt.} ``What is the largest factor of $54{,}321$ that is less than $54{,}321$? (A)~$6$, (B)~$121$, (C)~$18{,}107$, (D)~$54{,}320$.'' Gold answer: $\mathrm{C}$.

\noindent\textit{GRPO output (verbatim).}
\begin{quote}\small
To find the largest factor of 54,321 that is less than 54,321, we need to follow these steps:

\textbf{1. Identify the factors of 54,321:}
First, we need to find the prime factorization of 54,321. 54,321 is an odd number, so it is not divisible by 2. The sum of the digits of 54,321 is \(5 + 4 + 3 + 2 + 1 = 15\), which is divisible by 3. Therefore, 54,321 is divisible by 3. Dividing 54,321 by 3, we get \(54,321 \div 3 = 18,107\). Now, we need to check if 18,107 is a prime number. We do this by checking divisibility by prime numbers up to the square root of 18,107 (approximately 134.5). 18,107 is not divisible by 2, 3, 5, 7, 11, 13, 17, 19, 23, 29, 31, 37, 41, 43, 47, 53, 59, 61, 67, 71, 73, 79, 83, 89, or 97. Since 18,107 is not divisible by any of these primes, it is a prime number. Therefore, the prime factorization of 54,321 is \(3 \times 18,107\).

\textbf{2. Identify the largest factor less than 54,321:}
The largest factor of 54,321 is 18,107. To confirm, we check that 18,107 is less than 54,321. Indeed, \(18,107 < 54,321\).

Therefore, the largest factor of 54,321 that is less than 54,321 is \(\boxed{B}\).
\end{quote}

\noindent\textit{\textsc{SignBalance} output (verbatim).}
\begin{quote}\small
To find the largest factor of 54,321 that is less than 54,321, we need to follow these steps:

\textbf{1. Identify the largest factor less than 54,321}: The largest factor of a number is its smallest positive factor. Therefore, we need to find the largest factor of 54,321 that is less than 54,321.

\textbf{2. Check divisibility by small prime numbers}: We start by checking if 54,321 is divisible by the smallest prime number, 2. Since 54,321 is odd, it is not divisible by 2.

\textbf{3. Check divisibility by 3}: A number is divisible by 3 if the sum of its digits is divisible by 3. The sum of the digits of 54,321 is \(5 + 4 + 3 + 2 + 1 = 15\). Since 15 is divisible by 3, 54,321 is also divisible by 3. We can perform the division:
\[
54,321 \div 3 = 18,107
\]
So, $54{,}321 = 3 \times 18{,}107$.

\textbf{4. Verify that 18,107 is a factor}: We need to check if 18,107 is a factor of 54,321. We do this by performing the division:
\[
54,321 \div 18,107 = 3
\]
Since the division is exact, 18,107 is indeed a factor of 54,321.

\textbf{5. Conclusion}: The largest factor of 54,321 that is less than 54,321 is 18,107.

Therefore, the correct answer is \(\boxed{C}\).
\end{quote}

\paragraph{Example 2 (SAT-Math).}
\emph{Prompt.} ``There are $435$ voting members of the US House of Representatives. If $b$ voting members are in favor of a certain bill, which expression represents the percentage of the voting members in favor of the bill?'' Gold answer: $\mathrm{A}$.

\noindent\textit{GRPO output (verbatim).}
\begin{quote}\small
To find the percentage of voting members in favor of the bill, we need to express the number of voting members in favor as a fraction of the total number of voting members and then convert that fraction to a percentage.

The total number of voting members is 435. If \( b \) is the number of voting members in favor, then the number of voting members against is \( 435 - b \).

The percentage of voting members in favor is given by the formula:
\[
\text{Percentage} = \left( \frac{b}{435} \right) \times 100
\]

This can be rewritten as:
\[
\text{Percentage} = \frac{b \times 100}{435}
\]

So, the expression that represents the percentage of the voting members in favor of the bill is:
\[
\frac{b \times 100}{435}
\]

Therefore, the correct answer is:
\(\boxed{B}\)
\end{quote}

\noindent\textit{\textsc{SignBalance} output (verbatim).}
\begin{quote}\small
To find the percentage of voting members in favor of the bill, we need to express the number of voting members in favor as a fraction of the total number of voting members and then convert that fraction to a percentage.

The total number of voting members is 435. If \( b \) is the number of voting members in favor, then the number of voting members against is \( 435 - b \).

The percentage of voting members in favor is given by the formula:
\[
\text{Percentage} = \left( \frac{b}{435} \right) \times 100
\]

This can be rewritten as:
\[
\text{Percentage} = \frac{b \times 100}{435}
\]

Therefore, the expression that represents the percentage of the voting members in favor of the bill is:
\[
100 \left( \frac{b}{435} \right)
\]

So, the correct choice is:
\[
\boxed{A}
\]
\end{quote}

\paragraph{Interpretation.}
In both examples shown above, GRPO's reasoning content is correct, and the error appears only at the final letter-selection step. Under our analysis (Section~\ref{subsec:fail-cases}), a policy trained with the count-dependent GRPO advantage receives reinforcement on rollouts that produce the correct letter, regardless of whether that letter was reached through reasoning or by guessing among the candidates. Such a policy can therefore learn a letter-selection shortcut that does not faithfully follow the derivation. \textsc{SignBalance}, by decoupling the per-rollout magnitude from within-group composition, removes the gradient weight on rollouts that produce the correct letter by guessing. Accordingly, the resulting policy maps its derivations to letters more faithfully.

\section{Use of AI assistance}\label{app:ai-use}

We used an AI writing assistant to help polish the manuscript, including typo checking and refining word. All technical content, experimental design, results, and analyses are produced by the authors.

\end{document}